# TinyML Platforms Benchmarking

Anas Osman, Usman Abid, Luca Gemma, Matteo Perotto, and Davide Brunelli

Dept. of Industrial Engineering, University of Trento, I-38123 Povo, Italy
{name.surname}@unitn.it

**Abstract.** Recent advances in state-of-the-art ultra-low power embedded devices for machine learning (ML) have permitted a new class of products whose key features enable ML capabilities on microcontrollers with less than 1 mW power consumption (TinyML). TinyML provides a unique solution by aggregating and analyzing data at the edge on low-power embedded devices. However, we have only recently been able to run ML on microcontrollers, and the field is still in its infancy, which means that hardware, software, and research are changing extremely rapidly. Consequently, many TinyML frameworks have been developed for different platforms to facilitate the deployment of ML models and standardize the process. Therefore, in this paper, we focus on benchmarking two popular frameworks: Tensorflow Lite Micro (TFLM) on the Arduino Nano BLE and CUBE AI on the STM32-NucleoF401RE to provide a standardized framework selection criterion for specific applications.

## 1 Introduction

Machine Learning (ML) is at the forefront of innovation in technical and scientific applications, creating new insights on new and existing applications. Typically, running and analyzing large amounts of data on a complex ML algorithm requires a significant amount of resources and capabilities, which have been existing as barriers to the mainstreaming of ML in industry. However, with the improvement of powerful and energy-efficient embedded devices, ML inference is possible at the edge, and enables data analysis on the device as an alternative to the data exchange between servers and devices for decision making [1]. Initially, the field of edge ML focused on mobile inference, which ultimately led to several improvements for machine learning models such as quantization, sparsity, and pruning [2]. Recently, as IoT systems became mainstream, industry interest in extending Edge ML to microcontrollers grew to create a whole new potential for Edge ML as TinyML. The main goal of TinyML is to deploy ML models on ultra-low power devices to perform inference and achieve robust performance while breaking the power consumption barrier that has previously hindered such systems. TinyML eliminates the need for cloud server connectivity and improves responsiveness and privacy measures while running using a coin size battery.

Furthermore, the field is emerging and still in its infancy, with potential for innovative and state of the art applications to unlock its full potential [3]. Nonetheless, TinyML is already being implemented in many applications to provide smarter sensor technology that enables advanced monitoring to improve productivity and safety in many sectors. For example, predictive maintenance and monitoring of wind turbines is normally a cumbersome task as in most cases these turbines are located in remote areas and failures result in a long downtime. However, when predictive maintenance is implemented, downtime is significantly reduced, resulting in noticeable cost savings and an overall increase in quality and reliability. In [4], an Australian start-up company has developed a novel IoT device that can autonomously and simultaneously monitor the turbine during operation. The device is able to detect and report potential problems before they occur in the turbine's system. TinyML is also widely used in smart agriculture,as it is done for example by the Plant Village team, which developed an app that helps farms detect and treat potential diseases that affect crops [5]. In the health field, Solar Scare Mosquito focused on developing an IoT robotic platform that uses low-power, low-speed communication protocols to detect and warn of a potential mass breeding of mosquitoes [6].

The contribution of this paper is driven by the need to provide a standard framework and platform for TinyML use cases to build a foundation that drives the development of ML on edge devices. In particular, a comparison is made between two popular frameworks: Tensorflow Lite Micro (tested on an Arduino Nano BLE) and CUBE AI (tested on an STM32 NucleoF401RE) based on two TinyML applications.

This paper is structured as follows: Section II presents a summary overview of TinyML frameworks. In Section III, we provide a complete breakdown of benchmarking setting and tools implemented. Finally, the benchmarking is applied by comparing the two frameworks in Section IV and conclusions are drawn in Section V.

## 2 TinyML Frameworks

Due to the unlimited potential and great interest in TinyML to revolutionize various industries, many libraries and tools are constantly being developed and deployed to facilitate the implementation of ML algorithms on constrained platforms. TinyML frameworks can be divided into three different categories.

The most trivial approach refers to converting existing trained models to overcome MCU limitations. Thus, these tools typically use inference tools derived from well-known ML libraries such as TensorFlow [11], Scikit-Learn [12] or PyTorch [13], and port their code to run on devices with scarce resources.

The second category is based on the implementation of ML libraries specifically designed for MCUs to provide them with offline training and inference capabilities. It allows models to generate from data retrieved on the go from their device, which can immensely improve the model's accuracy and enables the implementation of unsupervised learning algorithms.

Finally, the last technique relies on the possibility of integrating a fully dedicated co-processor to support the main computing unit in ML-specific tasks. This strategy allows for an increase in computing power, although it is the least common approach because it significantly increases the price and complexity of processing platforms. Big tech companies are helping expand the TinyML ecosystem by contributing to open source development libraries [14]. Google has launched TensorFlow Lite for microcontrollers [15], which includes a set of tools to optimize TensorFlow models to make them run on mobile and embedded devices. The key to reducing the size and complexity of the framework is that it keeps only important features on the platform and eliminates less important ones. For instance, it fails to perform a full training of a model, but capable of making inferences on models that have already been trained on a cloud computing platform. This framework is based on two elements: the model converter, which converts TensorFlow models into optimized binary code that can be used on low-power MCUs, and the model interpreter, which executes the code generated by the converter.

The optimized models, which support a range of algorithms from the NN class, can run on several platforms, including smartphones, embedded Linux systems, and MCUs. In the case of MCUs, the optimized code is written in C++ and requires 32-bit processors. It has been successfully deployed on devices, such as the Arduino Nano, and other architectures, such as the ESP32 with ARM Cortex-M series processors. Given the prominence of Arduino, a special library for this platform is available through their IDE. STMicroelectronics is among the well-known electronics manufacturer that has developed specific libraries for its devices. Specifically, the STM32Cube.AI Toolkit [16] allows the integration of pre-trained NNs into STM32 ARM CortexM-based microcontrollers. It generates from the NN models provided by Tensorflow and Keras [17] STM32-compatible C code or from models in the standard ONNX format. As an interesting feature, STM32Cube.AI allows the execution of large NNs by storing weights and activation buffers in external flash memory or RAM. In addition, Microsoft has also contributed to the TinyML scene with the release of its open-source library Embedded Learning (ELL) [18].

This framework enables the design and deployment of pre-trained ML models on constrained platforms, such as ARM Cortex-A and Cortex-M based architectures like Arduino, Raspberry Pi and micro:bit. ELL acts as an optimizing cross-compiler that runs on a regular desktop computer and outputs C++ code that can be executed on the targeted single-board computer. The API of ELL can be used for both C++ and Python and uses pre-trained NN models provided by the Microsoft Cognition Toolkit (CNTK). The toolkit ARM-NN was introduced by ARM for integrating ML into their devices [19]. In addition to open source offerings, some institutions and companies have also launched privately licensed products. The Fraunhofer Institute for Microelectronic Circuits and Systems (IMS) has developed Artificial Intelligence Library for Embedded Systems (AIfES) running on even the smallest microcontrollers [20]. However, despite the variety of frameworks presented, they focus on only one type of ML algorithm, namely NN. Researchers and industry leaders have recently considered other ML techniques, such as decision trees, Naive Bayes classifier, k-Nearest Neighbors (k-NN), and others. For example, MicroML [21] is a novel technique that allows porting Support Vector Machine (SVM) and Relevance Vector Machine (RVM) algorithms to C code that can be used on a variety of MCUs, e.g. Arduino, ESP8266, ESP32 and others with C support. It supports the widely used scikit-learn toolkit and converts models generated by this library for use on 8-bit microcontrollers with 2 KB of RAM. A similar tool is m2cgen [22], which can transform the data from models formed with scikit-learn into native code, e.g. Python, C, Java. In this case, both the number of compatible algorithms and target programming languages is even larger than in m2cgen. Table I summarizes the main features of the frameworks considered in this section.

**Table 1.** Framework Comparison

| Framework | Algorithms | Compatible Platforms | Output Languages | External Libraries | Availability |
|---|---|---|---|---|---|
| TFLM | Neural networks | ARM Cortex-M | C++ 11 | Tensor Flow | Open Source |
| STM Cube AI | Neural networks | STM32 | C | Keras TensorFlow Lite Caffe ConvNetJs Lasagne | STM32 Devices only |
| ELL | Neural networks | ARM Cortex-M ARM Cortex-A | C/C++ | CNTK ARMDarknet ONNX | Open Source |
| ARM-NN AI | Neural networks | ARM Cortex-A ARM Mali ARM Ethos | C | TensorFlow Caffe ONNX | Open Source |
| AIfES | Neural networks | Raspberry Pi Windows (DLL) ARM Cortex-M4 | C | TensorFlow Keras | Private License |
| MicroMLGen | SVM RVM | Arduino ESP32 Arduino ESP8266 | C | Scikit-learn | Private License |
| m2cgen | Linear regression Logistic regression Neural networks SVM Decision tree | Multiple constrained nonconstrained platforms | Python C C # Java | Scikit-learn | Private License |

## 3 Benchmarking Setting

Machine learning benchmarks fall somewhere on the continuous sequence between low-level and application-level evaluation. Low-level benchmarks attempt target kernels at the core of many ML performance analysis, such as matrix multiplication, but hide critical elements such as memory bandwidth or model-level optimizations. Conversely, application-level benchmarks can hide the benchmark's goal behind other stages of the application pipeline. Our TinyML benchmark targets model inference and memory occupation. This section outlines the benchmarking setting for our two use cases. Each benchmark targets a specific use case with a different dataset, modelled on two separate targets. To perform this comparison with due diligence, we ensured that all parameters, device specifications, data used to train the model and model architecture were identical for both platforms.

### 3.1 Gesture Recognition Use Case

**Motivation** Gestures are expressive, meaningful body movements that involve physical movements of the fingers, hands, arms, or body to communicate and interact with the environment. New technology trends are driving the need to integrate such applications as part of smart systems to establish gesture recognition applications on tiny embedded devices.

**Dataset** As for the dataset, there are a number of open-source datasets relevant to TinyML use cases. However, we build a distinct dataset similar to the technique used by authors in [23], using the inertial measurement unit (IMU) via the LSM9DS1 sensor on the Arduino Nano 33 BLE. Figure 1 shows the spectral features extracted from the dataset of characters O, H, G, and C for acceleration data in the X, Y, and Z planes. From Figure 1 we can also identify the significant difference between the acquired data, which confirms the robustness of the dataset [23]. For each character of the 26 alphabet letters, 100 samples were acquired with a sampling frequency of 100 Hz, and each sample had an acquisition duration of 4000 ms, as shown in Figure 2.

**Network Architecture** For our application, we opted for a Convolutional Neural Network (CNN) to train our model compatible with TinyML deployment, as illustrated in Figure 3. The model is trained using Keras and Tensorflow Lite (TFL) libraries, which are compatible with both devices. In the case of the Arduino Nano, the model is converted to TFL using the Python API of the TFL converter. Then, our Keras model is written to disk in the form of a FlatBuffer, a special file format designed to save space [24-25].

**Model Optimization** There are two methods to choose from when optimizing a model: quantization and pruning. For the purpose of our application, we chose quantization. Quantization is still an active research topic, and there are many different options [26-27]. With dynamic quantization from float32 to int8, we were able to achieve promising results, as the model size was significantly reduced compared to the original version. In doing so, we managed to maintain reasonable accuracy when testing from 346KB to 275KB for TFLM and to 192KB for CUBE AI, while maintaining 85% accuracy.

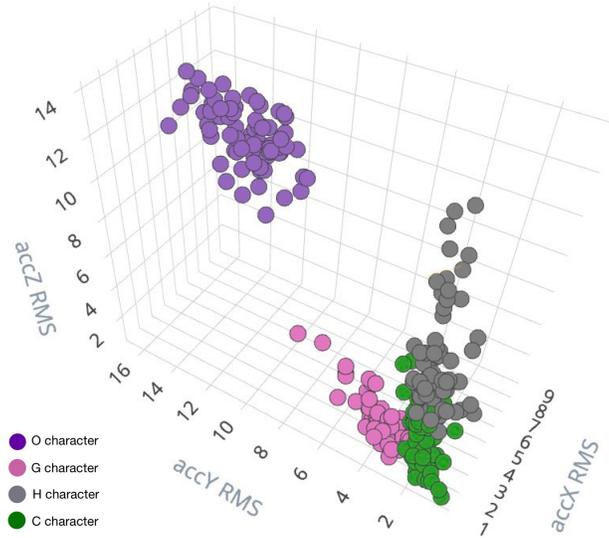

**Fig. 1.** Spectral Features of the dataset.

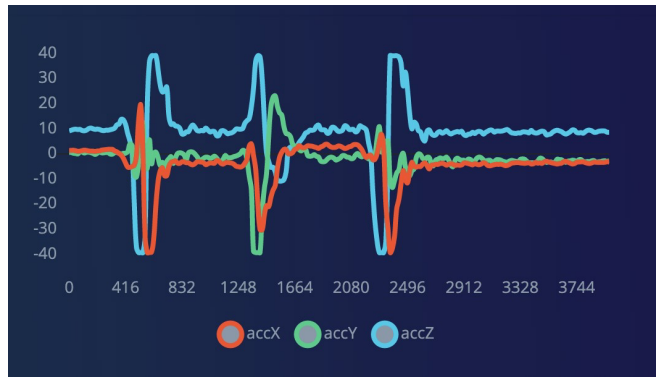

**Fig. 2.** N Character data representation.

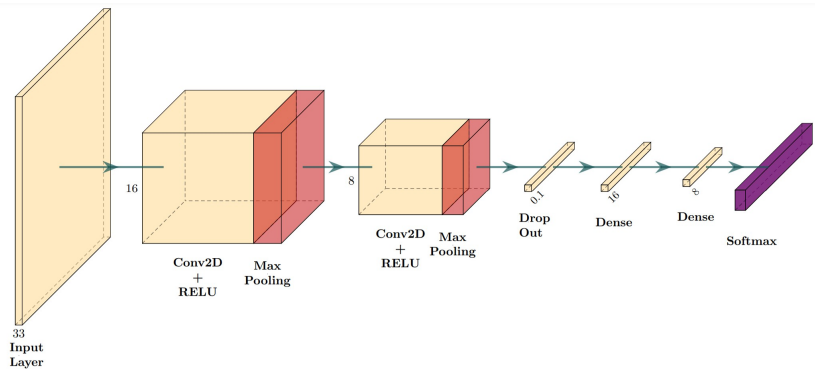

**Fig. 3.** Network Architecture

### 3.2 Wake Word Spotting Use Case

**Motivation** Wake Word Spotting (WWS), also known as Key Word Spotting, is a highlighted application and early use of TinyML, because voice command is an important aspect of human-machine interaction. The WWS application aims to run fully trained ML models on low-power devices to continuously monitor the environment for the wake-up word that triggers a particular functionality or service.

**Dataset** Dataset acquired from an open-source voice commands dataset featuring 500 speech samples with a duration of 1000 ms for 10 different wake-up words (UP, DOWN, YES, NO, GO, STOP, LEFT, RIGHT, ON, OFF) [28].

**Network Architecture** Similar to the Gesture Recognition application, we select a Convolutional Neural Network (CNN) to train our model compatible for TinyML deployment. The model is trained using Keras and Tensorflow Lite (TFL) libraries compatible with both devices.

**Model Optimization** Using the dynamic quantization feature from float32 to int8, we were able to achieve promising results as the model size was significantly decreased in comparison to the unoptimized size. The post training model was reduced from 650KB to 288KB for TLM and 247KB for X-Cube-AI.

### 3.3 Inference

After the model is converted, it is used on the two selected microcontrollers. For the Arduino platform, the C++ library for microcontrollers compatible with TFL is used to load the model and make predictions. The model is integrated as part of our applications shown in Figure 4 for the Gesture Recognition application to infer input data and display the output through the serial port. For the STM32 platform, the X-Cube AI supports both TFL and Keras model formats, allowing great flexibility in deploying the model on the microcontroller. We also generate C code using the platform-tools to represent and allocate all the resources of the model. Then, inference is performed based on given test data representing the characters and predictions are made based on the recognised character with the highest probability, which then indicates the accuracy of the model.

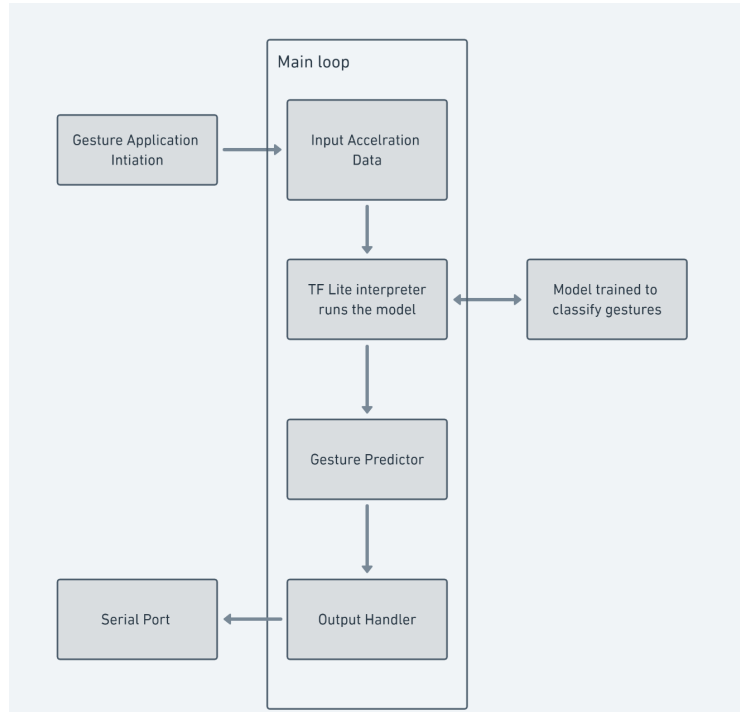

**Fig. 4.** Application Structure

## 4   TFLite-Micro vs X-CUBE-AI

For the comparison between the two frameworks, we chose two different microcontrollers that support the frameworks. The Arduino Nano 33 BLE and the STM NUCLEO F401RE were selected, as indicated in Table II. After successfully running and releasing our models on both platforms using the respective supported frameworks, as can be seen in Table III, the flash memory required by the application deployed on the X-CUBE-AI framework is significantly smaller than the same application deployed on the Arduino platform using the TFLM framework. Moreover, the inference time on the X-CUBE-AI framework is also significantly less in the first application but almost the same as TFLM for the second application.

Table 2. Comparison between the Two Devices

| Device | MCU | CPU | Clock | Memory | Framework |
|---|---|---|---|---|---|
| Arduino Nano BLE 33 | nRF52840 | 32-bit ARM Cortex M4 | 64MHz | 1MB | TFLM |
| STM32 NUCLEO-F401RE | LQFP64 | 32-bit ARM Cortex M4 | 84MHz | 512KB | X-CUBE-AI |

Table 3. Comparison between the Two Frameworks for The Two Use Cases

| Gesture Recognition Application | | | |
|---|---|---|---|
| Framework | Memory | DSP | Inference Time |
| TFLM | 275KB | 28ms | 30ms |
| X-CUBE-AI | 192KB | 5ms | 9ms |
| Wake Word Spotting Application | | | |
| Framework | Memory | DSP | Inference Time |
| TFLM | 288KB | 187ms | 193ms |
| X-CUBE-AI | 247KB | 162ms | 211ms |

## 5 Conclusion

Overall, the CUBE AI has a fairly straightforward system with a powerful interface that provides many tools for optimizing and handling the model and even generating code. On the other hand, the TFLM is more complex and requires many compromises to use the model successfully. Nevertheless, the TFLM outperforms the CUBE AI platform in terms of wide availability, because it is open-source and supports many devices. After running our two trained models on the two devices, the results show that CUBE AI performs better than the Tensorflow Lite Micro models, from the size differences to the more robust performance. However, it has the disadvantage of being supported only for STM devices and being software-oriented.

Finally, through the two discussed and compared applications in this paper, we conclude that the CUBE AI framework is better suited for memory-limited and performance-intensive TinyML applications. Future work would include further implementation of the two frameworks on more platforms through different performance demanding applications.